\documentclass[11pt]{article}

\usepackage[preprint]{acl}

\usepackage{times}
\usepackage{latexsym}

\usepackage[T1]{fontenc}

\usepackage[utf8]{inputenc}

\usepackage{microtype}

\usepackage{inconsolata}

\usepackage{graphicx}
\usepackage{subfigure}
\usepackage{booktabs}
\usepackage{tabularx}
\newcolumntype{L}{>{\raggedright\arraybackslash}X}
\newcolumntype{R}{>{\raggedleft\arraybackslash}X}
\usepackage{algorithm}
\usepackage{algorithmic}
\usepackage{amsmath}
\usepackage{multirow}
\usepackage[table]{xcolor}
\usepackage{amssymb}
\usepackage{makecell}
\usepackage{array}
\usepackage{enumitem}
\usepackage{tikz}
\usetikzlibrary{positioning, calc, arrows.meta}
\usepackage{float}
\usepackage{amsthm}

\usepackage{tcolorbox}
\tcbuselibrary{breakable, skins}
\newtcolorbox{promptbox}[2][]{%
  breakable,
  enhanced,
  colback=gray!4,
  colframe=gray!50,
  colbacktitle=gray!20,
  coltitle=black,
  fonttitle=\small\bfseries\sffamily,
  title={#2},
  boxrule=0.5pt,
  arc=3pt,
  left=6pt, right=6pt, top=4pt, bottom=4pt,
  #1
}

\usepackage{listings}
\usepackage{xcolor}

\lstdefinestyle{promptstyle}{
    basicstyle=\small\ttfamily,
    breaklines=true,
    frame=single,
    backgroundcolor=\color{gray!5},
    xleftmargin=10pt,
    xrightmargin=10pt,
    escapeinside={(*}{*)}, 
    showstringspaces=false
}

\title{DecoSearch: Complexity-Aware Routing and Plan-Level Repair for Text-to-SQL}

\author{
\textbf{Esteban Schafir}$^{1}$,
\textbf{Xu Zheng}$^{1}$,
\textbf{Hojat Allah Salehi}$^{1}$,
\\
\textbf{Zhuomin Chen}$^{1}$,
\textbf{Muhammad Yameen}$^{4}$,
\textbf{Mo Sha}$^{1}$,
\\
\textbf{Wei Cheng}$^{2}$,
\textbf{Dongsheng Luo}$^{3}$
\\
\\
$^{1}$Florida International University, $^{2}$NEC-Labs
\\
$^{3}$Singapore Management University, $^{4}$National University of Sciences and Technology Islamabad
}

\begin{document}
\maketitle

\begin{abstract}

Large Language Models (LLMs) have demonstrated remarkable capabilities in translating natural language to SQL, yet existing methods still falter on complex queries requiring multi-step, data-aware reasoning. We introduce DecoSearch, a training-free framework that addresses this by routing each query to the appropriate level of reasoning effort. A lightweight Schema Selector first prunes the full database schema to the relevant tables and columns. An LLM Judger then decides whether the question requires decomposition: straightforward questions follow a direct generation path and complex ones are escalated to a Directed Acyclic Graph (DAG) of atomic sub-questions, each solved by a targeted SQL generation step. A RAG component grounds the decomposer with semantically similar training examples.
DecoSearch achieves 70.53\% execution accuracy on BIRD and 88.31\% on Spider with a DeepSeek backbone, surpassing all training-free baselines. It also functions as a model-agnostic wrapper, consistently improving fine-tuned SQL generation backbones without any modification to the pipeline.

\end{abstract}

\section{Introduction}
\label{sec:Introduction}
The conversion of natural language questions into executable SQL (Text-to-SQL) is a fundamental challenge at the intersection of natural language understanding and database systems~\cite{katsogiannis2023survey,luo2025nl2sql,spider_dataset,bird_dataset}. The advent of Large Language Models (LLMs) has marked a new era of capability in this domain, enabling zero-shot or few-shot translation with impressive fluency~\cite{rajkumar2022evaluating,dail_sql,din,chess_sql}. However, the performance of these models degrades significantly when confronted with complex questions, those requiring nested queries, multi-step logical inference, or an awareness of the database's underlying content and structure~\cite{bird_dataset,li2024dawn}, or large databases with multiple tables that requires complex SQL queries with multiple JOIN operations~\cite{spider_dataset,li2024dawn}. For these problems, a single-pass, autoregressive generation process is often insufficient~\cite{scholak2021picard,din,chess_sql,chase_sql}, as it fails to explore the vast combinatorial space of possible SQL queries and cannot recover from early generation errors. 

Recent work has attacked this brittleness from several angles. Decomposition-based methods break a complex question into simpler sub-problems, improving the tractability of each step~\cite{din,l2m,eyal2023qpl,wang2025macsql,dea_sql}. Self-refinement approaches use execution feedback to iteratively debug generated SQL~\cite{selfrefine,critic,chen2024selfdebug,askari2025magic,excot}. Multi-candidate frameworks generate diverse SQL candidates and select the best one~\cite{sc,chase_sql,sun2023sqlpalm,lee2024mcssql,query_and_conquer}. Complexity-aware routing strategies allocate different pipelines to queries of different difficulty~\cite{elliesql,rethinking_agentic}. Each of these strategies addresses a real failure mode, but each operates at a single level of the problem. A self-debugging system can fix a syntactic error in a SQL query, but it cannot recognize that the underlying reasoning plan was flawed. A decomposition system can produce a sensible plan, but if one sub-problem turns out to be unsolvable as posed, it has no mechanism to revise its own plan. A multi-candidate system offers breadth, but it does not adaptively concentrate its computational budget on the most promising direction.


We observe that failures in complex Text-to-SQL generation are not monolithic. They arise from qualitatively different root causes that demand different interventions. Some failures are \emph{implementation-level}: the SQL contains a bug such as a wrong join condition or an incorrect filter, and an LLM debugger equipped with execution feedback can repair it directly. Other failures are \emph{strategy-level}: the chosen SQL formulation style (e.g., nested subqueries versus CTEs) is a poor fit for the problem, and a structurally different approach may succeed where iterative debugging of the current one does not. Still other failures are \emph{plan-level}: the high-level reasoning plan itself is flawed, a dependency is missing, or the decomposition is too coarse, and SQL-level repair alone may be insufficient. Existing methods typically focus on one or a subset of these failure modes. In particular, existing Text-to-SQL frameworks rarely explicitly diagnose the abstraction level at which a failure occurs and use that diagnosis to route the failure to a corresponding repair mechanism.

We introduce DecoSearch, a training-free framework that organizes Text-to-SQL generation as a multi-stage adaptive workflow with hierarchical failure routing. The framework employs a Judger-led escalation strategy: for every question, a specialized LLM judger evaluates whether the query can be solved directly or requires structural decomposition. For manageable questions, the system follows a Direct Path that generates and executes a single SQL query. For queries judged as complex, DecoSearch escalates to the plan level, decomposing the question into a Directed Acyclic Graph (DAG). Each sub-question is solved independently. When a node's SQL fails to execute or returns an empty result, an SQL Refiner first repairs the query using the execution feedback; failures that persist beyond this query-level repair trigger automatic topology refinement to restructure the reasoning plan. This ensures computational resources are spent proportionally to the depth of the failure.

Our primary contributions are as follows:

\begin{itemize}[leftmargin=*]
    \item We identify that failures in complex Text-to-SQL arise at distinct levels of abstraction (implementation, strategy, and planning) and introduce a hierarchical failure-routing architecture with a Judger-led direct path that efficiently resolves manageable queries while routing genuinely complex ones to structured decomposition.
    \item We introduce a two-level repair mechanism: an SQL Refiner that resolves implementation-level errors at individual DAG nodes, and a plan-level Topology Refiner that treats failures persisting beyond query-level repair as signs of a flawed decomposition, automatically restructuring the reasoning DAG in response. Evaluating on the BIRD and Spider benchmarks, DecoSearch achieves execution accuracies of 70.53\% and 88.31\%, respectively, and outperforms training-free baseline frameworks with a shared backbone.
\end{itemize}
\begin{figure*}[h!]
    \centering  
    \includegraphics[width=0.8\textwidth]{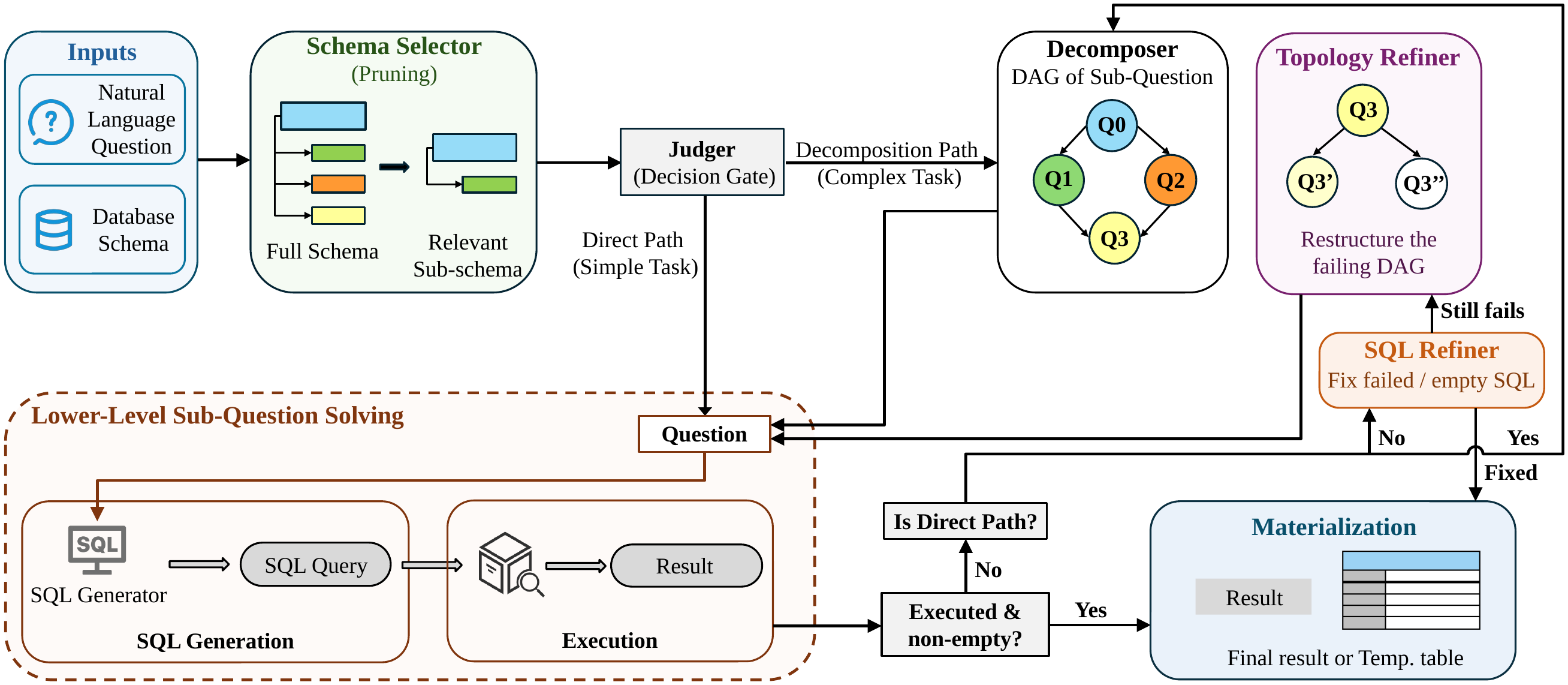}
    \caption{The pipeline begins with Pre-Processing via evidence-aware schema pruning. A Judger determines if the question is manageable; if so, the Direct Path generates and executes a single SQL query and exits if successful. Otherwise, the Decomposer produces a DAG of atomic sub-questions linked by \texttt{@[n]} dependency placeholders. At the Sub-Question Solving stage, each node is executed in topological order; upstream results are resolved as inline value lists for small result sets or materialized as temporary tables for large ones before the downstream SQL is generated. Failed or empty node queries are first repaired by the SQL Refiner; persistent failures trigger the Topology Refiner to rewrite the DAG and retry.}
    \label{fig:pipeline_diagram}
\end{figure*}
\section{Related Work}
\label{sec:related_work}

\noindent \textbf{\textit{Text-to-SQL with LLMs.}}
Early Text-to-SQL systems relied on semantic parsing and schema-aware neural encoders~\cite{ratsql,igsql}. The rise of large-scale LLMs shifted the paradigm toward in-context learning, enabling strong zero-shot and few-shot performance~\cite{rajkumar2022evaluating,dail_sql}. However, single-pass generation remains brittle on complex queries requiring multi-table joins or data-aware reasoning~\cite{bird_dataset,li2024dawn}, a challenge made even more acute by newer, harder benchmarks such as Spider~2.0~\cite{spider2}, which features enterprise-scale databases where state-of-the-art models achieve below 25\% accuracy.

\noindent \textbf{\textit{Decomposition-based Methods.}}
A natural response to this brittleness is to decompose complex questions into simpler sub-problems~\cite{l2m,din,eyal2023qpl,wang2025macsql}. DIN-SQL~\cite{din} pioneered decomposed in-context learning with self-correction for Text-to-SQL. DEA-SQL~\cite{dea_sql} introduces a workflow-based paradigm that counters attention diffusion in monolithic prompting through sequential filtering, classification, and correction stages. Our work extends these ideas by formalizing the decomposition as a dependency DAG and enabling plan-level repair when sub-problems fail.

\noindent \textbf{\textit{Schema Linking.}}
Passing the full database schema to every LLM call degrades generation quality and inflates cost. CHESS~\cite{chess_sql} integrates LLM-driven schema pruning as a dedicated agent. RSL-SQL~\cite{rsl_sql} further improves schema linking with bidirectional pruning, achieving high recall while cutting input columns substantially. DecoSearch's schema selector follows this paradigm, applying LLM-driven keyword extraction and two-stage table and column selection before any generation call.

\noindent \textbf{\textit{Self-Refinement and Execution Feedback.}}
Several methods improve robustness by iteratively revising SQL using execution feedback~\cite{selfrefine,critic,chen2024selfdebug,askari2025magic}. ExCoT~\cite{excot} extends this with execution-supervised fine-tuning, iteratively improving open-source models using only SQL execution accuracy as the reward signal. These approaches address implementation-level failures but cannot restructure an incorrect reasoning plan. DecoSearch adopts execution-guided repair at the node level through its SQL Refiner, and complements it at the plan level: when a failure persists beyond query-level repair, the system rewrites the DAG rather than continuing to patch individual queries.

\noindent \textbf{\textit{Multi-Candidate, Routing, and Test-Time Scaling.}}
CHASE-SQL~\cite{chase_sql} generates diverse candidates via multiple CoT strategies and selects the best via a fine-tuned pairwise ranker. Agentar-Scale-SQL~\cite{agentar_scale_sql} combines RL, refinement, and parallel synthesis at the query level. CHESS~\cite{chess_sql} and MCS-SQL~\cite{lee2024mcssql} use self-consistency voting over multiple candidates; Query and Conquer~\cite{query_and_conquer} shows that execution-guided selection alone enables smaller models to match much larger ones. Complexity-aware routing has also emerged as a practical tool: EllieSQL~\cite{elliesql} routes queries to different generation pipelines based on estimated difficulty, achieving substantial token savings, while Rethinking Agentic Workflows~\cite{rethinking_agentic} benchmarks test-time scaling strategies for Text-to-SQL and finds that routing and decomposition consistently outperform naive candidate scaling. DecoSearch uses a Judger-led routing strategy to select between direct generation and hierarchical decomposition, reserving the structured DAG approach for queries requiring deeper reasoning.
\section{Methodology}
\label{sec:methodology}
\subsection{Problem Formulation}
Given a natural language question $Q$, a database $\mathcal{D}$ with schema $\mathcal{S}$, and optionally a set of domain hints (``evidence''), our goal is to produce an SQL query $\hat{Q}$ such that $\text{Execute}(\hat{Q}, \mathcal{D}) = \text{Execute}(Q^*, \mathcal{D})$, where $Q^*$ is the ground-truth query. We evaluate correctness via execution accuracy (EX): a query is correct if and only if its result set semantically matches the gold result set.

\subsection{Overview}
DecoSearch treats Text-to-SQL generation as an adaptive, multi-level reasoning process. Rather than committing every question to the same generation procedure, the system first assesses query complexity and routes each question to the cheapest path that is likely to succeed. For straightforward questions, a direct generation attempt is sufficient. For complex questions, those requiring multi-table joins, nested logic, or multi-step inference; the system escalates to a structured decomposition that breaks the question into a DAG of simpler, independently solvable sub-questions. Failures are diagnosed and repaired at the appropriate level of abstraction: when a node's SQL fails to execute or returns an empty result, an SQL Refiner first repairs the query using the execution feedback; if the failure persists, the system treats it as a sign of a flawed reasoning plan and triggers topology refinement of the DAG. The entire framework is training-free and operates at test-time, relying solely on the reasoning capabilities of an off-the-shelf LLM.

Our proposed framework consists of six primary components: (1) a Schema Selector that prunes the database schema to the most relevant tables before any LLM call, (2) a Judger-Led Escalation strategy that determines the appropriate reasoning depth, (3) a Direct Path for manageable questions, (4) a Query-Level SQL Generation step for sub-problem solving, (5) an SQL Refiner for query-level repair, and (6) a Hierarchical Control Loop for dynamic topology refinement. This tiered approach ensures that computational resources are allocated proportionally to the complexity of the question, with decomposition reserved for queries the Judger identifies as complex. The full pipeline is given in Figure ~\ref{fig:pipeline_diagram}. We describe each component below.

\subsection{Schema Selection}
\label{sec:schema_selector}
Real-world databases often contain dozens of tables and hundreds of columns, the vast majority of which are irrelevant to any given question.
Passing the full schema to every LLM call inflates
the prompt and, more critically, degrades generation quality by filling the context with distracting schema noise.
DecoSearch therefore applies a lightweight schema pruning step before any LLM call in the pipeline.

Our Schema Selector follows the schema-pruning approach introduced in CHESS~\cite{chess_sql} and DeepEye-SQL~\cite{deepeye_sql}, and is consistent with the bidirectional pruning principles of RSL-SQL~\cite{rsl_sql}, adapting their keyword-based filtering strategy to our pipeline.
The Schema Selector operates in two stages.
First, it issues a single LLM call to extract a set of task-relevant keywords and entities from the NLQ and the evidence string (e.g., column names, value filters, domain terms).
Second, it uses those keywords to drive an LLM-based table selection call, followed by a column selection call within the chosen tables.
Only the selected tables and their columns are serialized into the schema string passed to the Judger, Decomposer, SQL Generator, SQL Refiner, and Topology Refiner.

This pruning reduces prompt length by up to an order of magnitude for large databases and focuses the LLM's attention on relevant schema elements.

\subsection{Judger-Led Escalation and Direct Path}
The pipeline begins with a specialized LLM Judger that evaluates the complexity of the natural language question (NLQ) relative to the database schema. The Judger is a single zero-shot LLM call that receives two inputs: (1) the NLQ and (2) the pruned database schema produced by the preceding schema-selection step. It outputs a structured JSON object of the form \texttt{\{"needs\_decomposition": true/false, "reasoning": "..."\}}, providing both a binary routing decision and a brief rationale.

The Judger prefers the direct path by default, escalating to decomposition only for questions with conflicting aggregation grains, extreme join complexity, multi-phase reasoning where step two depends on a materialized result from step one, or evidence-driven formulas that would make a single query excessively nested. This routing philosophy is consistent with recent work on complexity-aware allocation~\cite{elliesql,rethinking_agentic}, which shows that matching computational effort to query difficulty yields significant accuracy and efficiency gains over uniform strategies.

When decomposition is not needed, the pipeline proceeds directly to the Direct Path described below; when it is, the system skips the direct path and proceeds to DAG-based decomposition.

For questions judged as manageable, the system follows a Direct Path: a single SQL query is generated using a zero-temperature LLM call and executed against the database. If execution succeeds, the result is returned immediately, keeping both latency and cost low. If execution fails, the framework falls through to the decomposition stage, which systematically breaks the original query into a sequence of simpler, more verifiable sub-tasks.

\subsection{Plan-Level Search: Task Decomposition}
When the Judger mandates escalation or the direct path fails to execute, the system initiates an upper-level plan by decomposing the complex NLQ into a DAG. In this graph, nodes represent simpler, atomic sub-questions, and directed edges represent data dependencies between them. The topological order of the graph determines the execution sequence, ensuring that each node's inputs are fully materialized before it executes.

\textbf{Example 1.} Consider the following question from the BIRD benchmark:
\begin{quote}
\textit{``List all patients who were followed up at the outpatient clinic who underwent a laboratory test in October 1991 and had a total blood bilirubin level within the normal range.''}
\end{quote}

\begin{figure}
    \centering
    \includegraphics[width=0.7\linewidth]{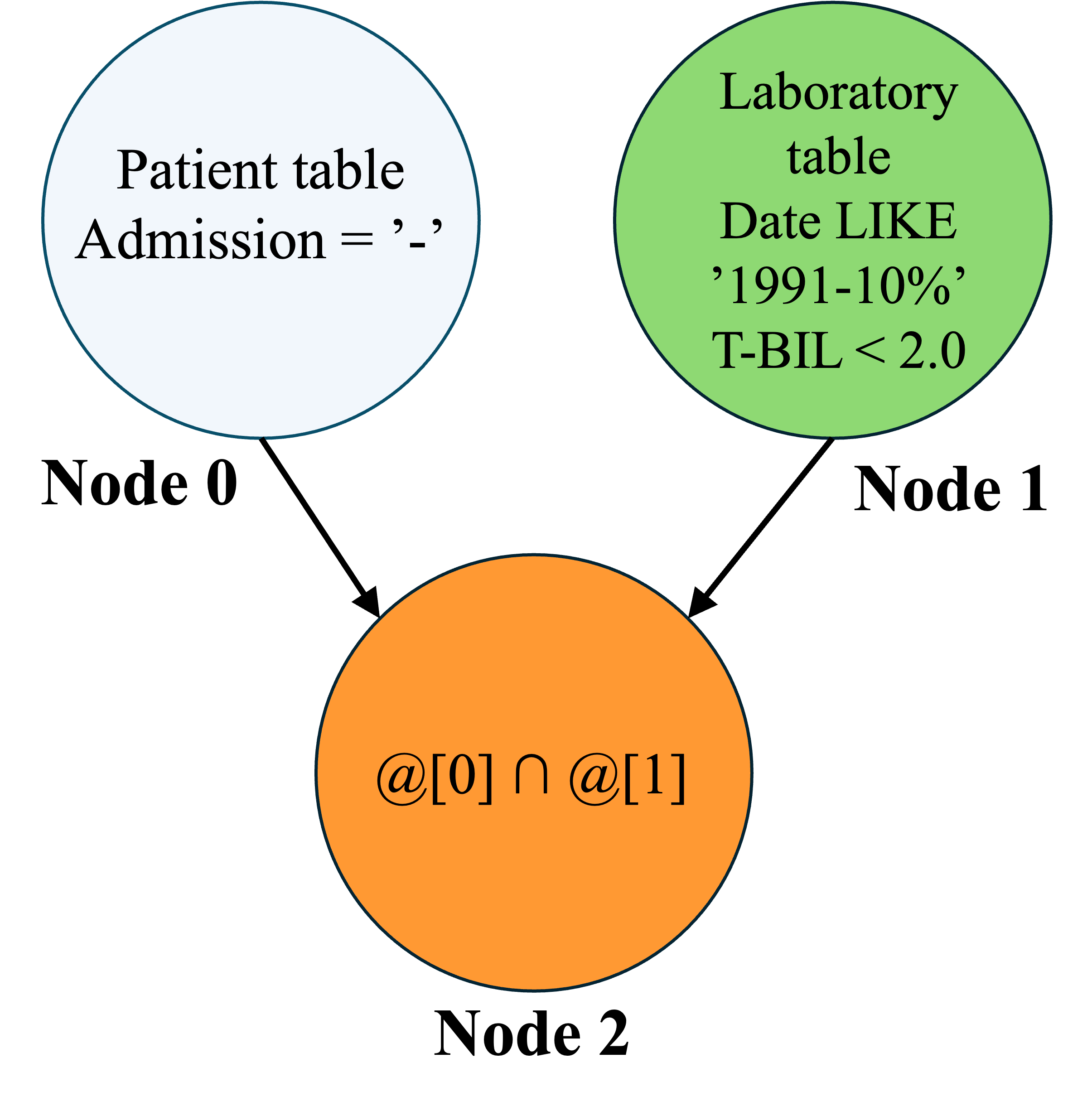}
    \caption{DAG for the \texttt{thrombosis\_prediction} example. Node~2 executes last, injecting results of Nodes~0 and~1 via \texttt{@[n]} placeholders. }
    \label{fig:dag_example}
\end{figure}

The Decomposer breaks this into the three-node DAG shown in Figure~\ref{fig:dag_example}. Nodes 0 and 1 are independent and can be solved in parallel; Node 2 consumes both their results. The \texttt{@[n]} syntax is a dependency placeholder: at execution time, the result set of Node $n$ is resolved and injected into the downstream node's prompt and SQL before generation. For small result sets ($\leq$100 rows), values are inlined as a SQL value list (e.g., an \texttt{IN (\ldots)} literal); for larger result sets, the upstream result is materialized into a SQLite temporary table (\texttt{temp\_ds\_node\_n}), which the downstream node can query by name. This design prevents the LLM from having to express the full multi-table logic in a single query while remaining efficient regardless of result cardinality (see Appendix~\ref{app:placeholder_resolution} for details).



To improve the logical soundness of this initial decomposition, we employ a Retrieval-Augmented Generation (RAG) strategy. Before generating the graph, the system retrieves a set of successful question-decomposition pairs from a pre-populated knowledge base. These examples are dynamically injected into the decomposer's prompt as few-shot demonstrations, grounding the LLM in relevant, structured examples and guiding it toward more coherent and viable task graphs.

The knowledge base is constructed offline from prior successful runs on the benchmark's training split: experiment logs are parsed to extract question–sub-question pairs for which the pipeline produced a correct final answer, and these are stored as a JSONL file. At inference time, we encode the incoming question using a sentence embedding model~\cite{reimers-2019-sentence-bert} and retrieve the $k=3$ most similar training examples by cosine similarity.

\subsection{Query-Level SQL Generation}
\label{sec:generation}
For each node in the decomposition graph, the system generates a single SQL query using a zero-temperature LLM call (full prompt in Appendix~\ref{app:prompt_generator}). The prompt instructs the model to produce a valid SQL query grounded in the pruned schema and any available evidence hints, while applying standard robustness conventions such as case-insensitive string comparisons. The generated query is executed against the database; if it succeeds, its result is materialized and passed to downstream nodes via the \texttt{@[n]} placeholder mechanism. If execution fails or returns an empty result, the query is passed to the SQL Refiner. The full pipeline is described in Algorithm~\ref{alg:sc_algorithm}.

\subsection{Query-Level Repair: SQL Refiner}
DecoSearch first attempts to correct a failing query at the node level before revising the reasoning plan, resolving implementation-level errors without discarding the decomposition.
Query-level repair triggers when a node's generated SQL fails to execute or returns an empty result, indicating a likely implementation error, e.g., a misnamed column, an incorrect join condition, or an overly restrictive filter, that does not require modifying the sub-question itself.
On failure, the SQL Refiner prompts the LLM with the sub-question, the pruned schema, the evidence, and the history of prior attempts with their execution errors (full prompt in Appendix C.4). The LLM produces a corrected query, which is executed and, if successful, materialized and passed to downstream nodes. Refinement is limited to one attempt per node; a query that still fails signals failure to the Hierarchical Control loop.

\subsection{Hierarchical Control: Dynamic Topology Refinement}
To improve resilience against flawed initial decompositions, DecoSearch maintains a hierarchical feedback loop that enables dynamic plan correction at the graph level. This mechanism allows the system to recover from cases where the decomposition strategy itself is the root cause of failure, rather than individual SQL generation.
Topology refinement is triggered when a node's SQL still fails to execute, or still returns an empty result, after query-level repair by the SQL Refiner. A failure that persists beyond query-level repair signals that the sub-question may be ill-formed, overly broad, or structurally inconsistent with the available schema; problems that cannot be resolved by regenerating or refining a query for the same sub-question.
Upon receiving the failure signal, the framework activates the Topology Refiner module that provides the LLM with the current DAG structure, the failing node's sub-question, and the execution error from the failed query. The LLM's task is to produce a revised DAG, which may involve re-partitioning dependencies, inserting intermediate nodes to break down an overly complex sub-question, or restructuring the data-flow to better align with the schema. The pipeline then adopts the revised graph and resumes execution from the first modified node, preserving the results of all unaffected upstream nodes. Refinement is bounded by a budget $B$ (the maximum number of topology refinement attempts per query). If the revised DAG also fails to execute and the budget is exhausted, the system accepts the best available result for the failing node and continues execution of all downstream nodes. This graceful degradation ensures that a single irresolvable sub-problem does not abort the entire pipeline.

\begin{algorithm}[hb!]
    \small
    \caption{DecoSearch Pipeline}
    \label{alg:sc_algorithm}
    \begin{algorithmic}[1]
    \REQUIRE Natural Language Question $Q$, Database $\mathcal{D}$, Schema $\mathcal{S}$, topology-refinement budget $B$
    \ENSURE Final SQL Result $\mathcal{R}$

    \STATE \COMMENT{\textbf{Step 1: Schema Selection}}
    \STATE $\mathcal{S}_{pruned} \leftarrow \text{SchemaSelector}(Q, \mathcal{S})$

    \STATE \COMMENT{\textbf{Step 2: Judger --- Direct Path or Decompose}}
    \IF{\text{Judger}($Q, \mathcal{S}_{pruned}$) is \text{Direct Path}}
        \STATE $\hat{Q} \leftarrow \text{SQLGenerator}(Q, \mathcal{S}_{pruned})$
        \STATE $R, \text{Error} \leftarrow \text{Execute}(\mathcal{D}, \hat{Q})$
        \IF{$\text{Error} = \text{NONE}$}
            \RETURN $R$ \COMMENT{Direct execution succeeded}
        \ENDIF
        \STATE \COMMENT{Execution failed: fall through to decomposition}
    \ENDIF

    \STATE \COMMENT{\textbf{Step 3: Decomposition}}
    \STATE $G \leftarrow \text{Decomposer}(Q, \mathcal{S}_{pruned})$ \COMMENT{DAG of sub-questions}
    \STATE $Queue \leftarrow \text{TopologicalSort}(G)$
    \STATE $Results \leftarrow \{\}$

    \STATE \COMMENT{\textbf{Step 4: Hierarchical Control \& Materialization}}
    \WHILE{$Queue$ is not empty}
        \STATE $n \leftarrow Queue.\text{pop}(0)$
        \STATE $Results[n], G \leftarrow \text{SolveNode}(n, G, Results, \mathcal{S}_{pruned}, \mathcal{D}, B)$ \COMMENT{Alg.~\ref{alg:solve_node}, App.~\ref{app:solve_node}}
        \IF{$G \text{ is revised}$}
            \STATE $Queue \leftarrow \text{TopologicalSort}(G)$
        \ENDIF
    \ENDWHILE

    \RETURN $Results[\text{final}]$
    \end{algorithmic}
    \end{algorithm}

\section{Experiments}
\label{sec:experiments}
We conduct empirical evaluations to assess the performance of our proposed framework. The experiments are designed to answer three primary research questions: (1) How does our hierarchical search framework compare against state-of-the-art (SOTA) Text-to-SQL models, including under a shared backbone? (2) How do Judger-led routing and the two repair mechanisms contribute to the final accuracy? (3) Can the proposed method be combined with fine-tuned models to improve accuracy?

\paragraph{Datasets.}
We evaluate on two standard benchmarks. Spider~\cite{spider_dataset} contains 2,147 testing queries across 200 databases. It emphasizes structural SQL complexity (multi-table joins, nested queries). BIRD~\cite{bird_dataset} contains 1,534 development queries across 11 databases designed to test ``data-aware'' reasoning, requiring the model to understand actual database content (e.g., value distributions, cell-level patterns).

\subsection{Experimental Setup}
\noindent \textbf{Baselines.} Our method is tested with two categories of baselines: Training-Free and Fine-tuned methods. Training-free methods operate solely via prompting without any task-specific fine-tuning, including CHESS~\cite{chess_sql}, GenaSQL~\cite{genasql}, and DAIL-SQL~\cite{dail_sql}. Fine-tuned methods additionally train or adapt model weights on Text-to-SQL data, including XiYan-SQL~\cite{xiyan_sql}, Agentar-Scale-SQL~\cite{agentar_scale_sql}, CHASE-SQL~\cite{chase_sql}, and OmniSQL~\cite{omnisql}. Because our approach focuses on test-time reasoning strategy, query routing, structured decomposition, and plan-level repair, rather than on model parameters or training data, it is orthogonal to these fine-tuned methods and can be readily combined with them to yield further improvements.

\noindent \textbf{Implementation Details.} Our framework is model-agnostic. Unless otherwise stated, all LLM-driven components (Judger, Decomposer, SQL Generator, SQL Refiner, Topology Refiner) use DeepSeek as the backbone model with a single greedy generation per node. For the shared-backbone comparison (\S\ref{sec:results}) and the routing and repair analysis (\S\ref{sec:ablation}), DecoSearch and the re-run baselines use gpt-5.4-mini. The SQL Refiner is limited to a single refinement attempt per node. 
SQL Generator prompts are lightly adapted per dataset, with an identical core reasoning structure. Full hyperparameter settings are reported in Appendix~\ref{app:hyperparams}.

\begin{figure}[h!]
  \centering
  \includegraphics[width=\linewidth]{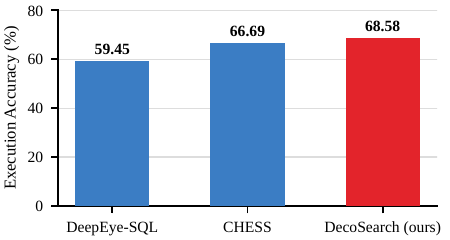}
  \caption{Execution Accuracy on BIRD dev with gpt-5.4-mini as main backbone.}
  \label{fig:shared_backbone}
\end{figure}

\noindent \textbf{Evaluation Metric.} The primary metric is Execution Accuracy. A generated SQL query is deemed correct if and only if its execution result set is identical to that of the ground-truth query. Result sets are canonicalized to be robust against arbitrary row and column ordering, as well as known surface-level formatting artifacts (e.g., concatenated vs.\ separate string tokens that convey the same information).

\begin{figure*}[!t]
  \centering
  \begin{minipage}[t]{0.45\textwidth}
    \centering
    \includegraphics[width=\linewidth]{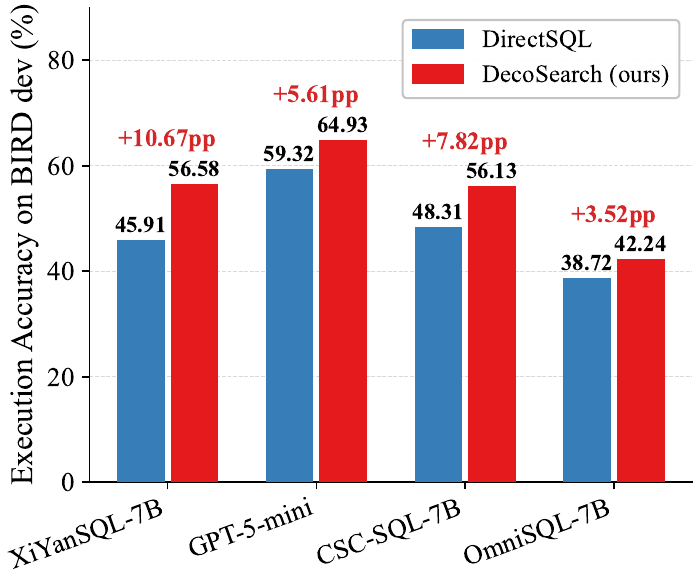}
    \caption*{(a) Fine-tuned backbones}
  \end{minipage}
  \hfill
  \begin{minipage}[t]{0.45\textwidth}
    \centering
    \includegraphics[width=\linewidth]{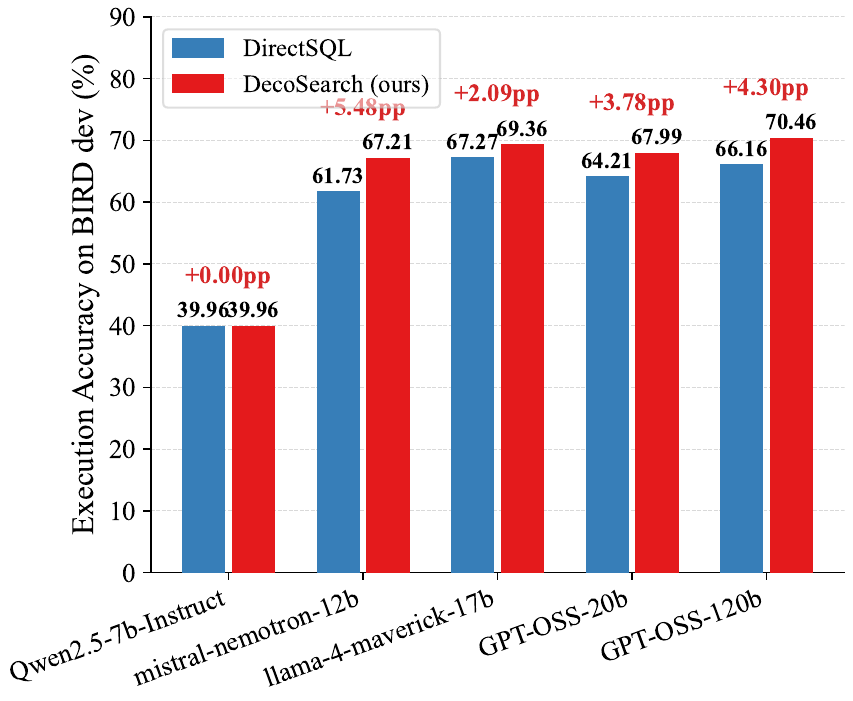}
    \caption*{(b) Open-source backbones}
  \end{minipage}
  \caption{Execution Accuracy on BIRD when pairing DecoSearch with fine-tuned (a) and open-source (b) backbone models, compared against their DirectSQL baseline. DecoSearch consistently improves over the direct generation baseline regardless of the backbone, demonstrating its model-agnostic nature.}
  \label{fig:finetune_comparison}
\end{figure*}

\begin{figure*}[!t]
  \centering
  \includegraphics[width=\textwidth]{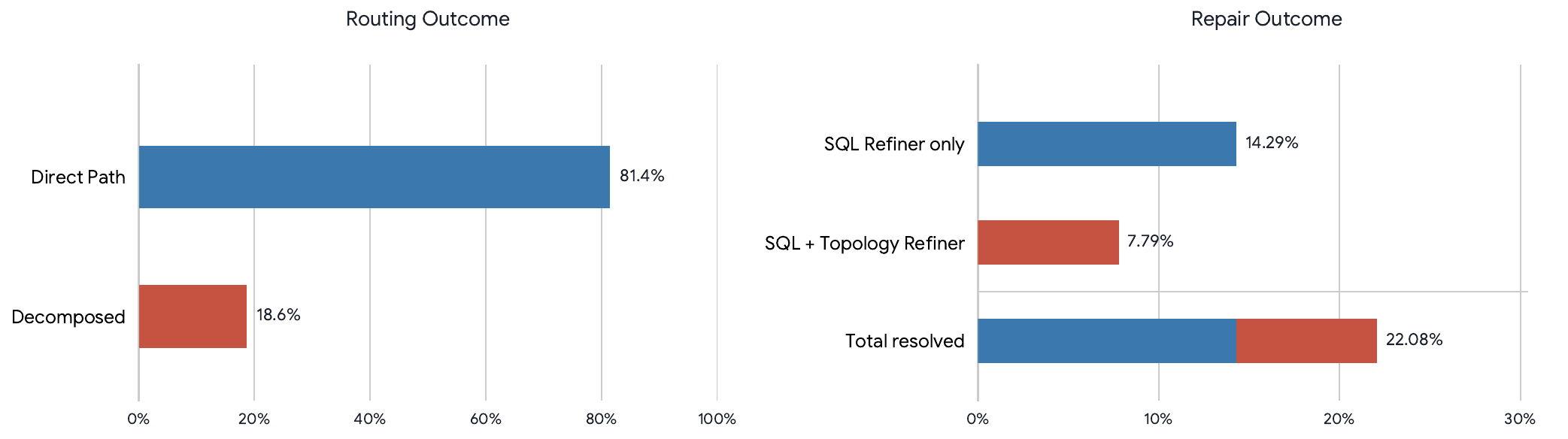}
  \caption{Routing and repair outcome of DecoSearch on BIRD. (a) Routing outcome: share of questions handled by the direct path (81.4\%) versus decomposition (18.6\%). (b) Repair outcome: share of questions resolved by the SQL Refiner only (14.29\%) and by the SQL Refiner followed by the Topology Refiner (7.79\%), together 22.08\% resolved in total.}
  \label{fig:breakdown}
\end{figure*}

\subsection{Results and Analysis}
\label{sec:results}

\paragraph{Overall Performance.}
Evaluation results are summarized in Table~\ref{tab:full_comparison}. On BIRD and Spider, DecoSearch achieves 70.53\% and 88.31\% EX, respectively, consistently outperforming all training-free baselines. These findings validate that our primary performance gains stem from the hierarchical decomposition and the Judger-led routing strategy. For reproducibility, we also verify that this advantage is not an artifact of the backbone LLM: re-running CHESS and DeepEye-SQL alongside DecoSearch on the full BIRD development set with a shared gpt-5.4-mini backbone (Figure~\ref{fig:shared_backbone}), DecoSearch still leads, at 68.58\% EX, outperforming CHESS by 1.89 pp and DeepEye-SQL by 9.13 pp.

\begin{table}[h!]
\centering
\caption{Comprehensive comparison of Text-to-SQL frameworks on BIRD and Spider benchmarks, categorized by learning approach. ``Model Avail.'' indicates whether model weights are publicly available.}
\label{tab:full_comparison}
\footnotesize
\setlength{\tabcolsep}{3.4pt}
\begin{tabular}{lccc}
\toprule
\textbf{Method} & \textbf{Model} & \textbf{BIRD dev} & \textbf{Spider dev} \\
 & \textbf{Avail.} & \textbf{EX (\%)} & \textbf{EX (\%)} \\
\midrule
\multicolumn{4}{l}{\textit{Training-free}} \\
DeepEye-SQL & $\times$ & 68.51 & 84.44 \\
GenaSQL     & $\times$ & 65.32 & 81.93 \\
CHESS       & $\times$ & 69.60 & 87.33 \\
DAIL-SQL    & $\times$ & 59.39 & 83.56 \\
DIN-SQL     & $\times$ & 39.57 & 64.51 \\
\midrule
\multicolumn{4}{l}{\textit{Open-source}} \\
GPT-OSS-120b         & $\checkmark$ & 66.16 & 84.16 \\
GPT-OSS-20b          & $\checkmark$ & 64.21 & 83.09 \\
Llama-4-Maverick-17B & $\checkmark$ & 67.27 & 85.42 \\
Mistral-nemotron-12B & $\checkmark$ & 61.73 & 81.93 \\
Qwen2.5-7B           & $\checkmark$ & 39.96 & 77.27 \\
DeepSeek             & --           & 66.81 & 85.56 \\
\midrule
\textbf{DecoSearch (ours)} & -- & \textbf{70.53} & \textbf{88.31} \\
\bottomrule
\end{tabular}
\end{table}

\paragraph{DecoSearch as a General-Purpose Wrapper for Fine-Tuned Models.}
A key property of DecoSearch is that its hierarchical search framework is entirely model-agnostic: the backbone LLM can be swapped without any modification to the pipeline. To demonstrate this, we pair DecoSearch with several SOTA fine-tuned SQL generation models and evaluate on BIRD. Figure~\ref{fig:finetune_comparison} shows that DecoSearch consistently improves over the corresponding DirectSQL baseline for every backbone, with gains of $+10.67$ pp (XiYanSQL-7B), $+5.61$ pp (GPT-5-mini), and $+3.52$ pp (OmniSQL-7B). These results confirm that DecoSearch's decomposition and routing strategy is complementary to model-level fine-tuning: it provides an additional layer of structured reasoning on top of whatever SQL generation capability the backbone already possesses, and can therefore be applied as a drop-in enhancement to any existing fine-tuned system.

\subsection{Ablation Study}
\label{sec:ablation}
To quantify the contribution of each component, we conduct an ablation study on a 770-query subset of BIRD, using DeepSeek as the backbone, shown in Table~\ref{tab:ablation}. The Judger is the dominant component: forcing all queries through decomposition causes the largest drop ($-10.65$ pp), revealing that DAG decomposition actively hurts simple queries. The Judger's role is to prevent from over-engineering. The schema selector is the second most impactful component ($-1.82$ pp). Removing decomposition entirely costs only $-1.43$ pp, confirming it targets a small but consequential subset. Topology refinement provides an additional incremental gain ($-0.78$ pp), with the primary benefit coming from routing and hierarchical decomposition.

Figure~\ref{fig:breakdown} supports this picture at the level of individual routing and repair decisions on the full BIRD development run. Direct-path questions are solved at a much higher rate than decomposed ones, consistent with the Judger reserving decomposition for genuinely harder questions rather than diluting easy ones. Among questions that need repair, most are resolved by the SQL Refiner alone, at roughly twice the success rate of those severe enough to also require topology refinement.

\begin{table}[!t]
\centering
\caption{Ablation study on BIRD (dev). Each row removes one component from the full DecoSearch pipeline. $\Delta$ EX is the absolute change in Execution Accuracy relative to the full system.}
\label{tab:ablation}
\small
\setlength{\tabcolsep}{4pt}
\begin{tabular}{lrr}
\toprule
\textbf{Configuration} & \textbf{EX (\%)} & \textbf{$\Delta$ EX} \\
\midrule
Full Pipeline                        & 69.48 & ---      \\
\midrule
w/o Judger (always decompose)        & 58.83 & $-$10.65 \\
w/o Decomposition (direct path only) & 68.05 & $-$1.43  \\
w/o Topology Refinement              & 68.70 & $-$0.78  \\
w/o Schema Selector                  & 67.66 & $-$1.82  \\
\bottomrule
\end{tabular}
\end{table}

\section{Conclusion}
\label{sec:conclusion}

We presented DecoSearch, a training-free framework that adaptively matches computational effort to query difficulty by routing each question through a Judger-Led Escalation strategy: simple queries are resolved via direct generation, while complex ones are escalated to a DAG decomposition process with query-level repair by an SQL Refiner and automatic Topology Refinement. DecoSearch achieves 70.53\% execution accuracy on BIRD and 88.31\% on Spider, surpassing all training-free baselines. The ablation study confirms that Judger-led routing is the dominant contributor ($-10.65$ pp when disabled), underscoring that DecoSearch's value lies in the principled orchestration of routing, decomposition, and two-level repair.


\section{Limitations}
DecoSearch relies on executable database feedback during inference. The SQL Refiner and the topology refiner are triggered by SQL execution errors or empty results at individual DAG nodes, which assumes that queries can be executed against the target database and that execution failures are informative signals of plan level flaws. This assumption is common across most training free Text-to-SQL methods that incorporate self-refinement or candidate selection, since execution feedback is the strongest unsupervised signal available at test time. For settings where execution is unavailable, prohibitively expensive, or returns semantically wrong but syntactically valid results, DecoSearch, like other execution guided methods, cannot intervene. Nevertheless, the framework remains applicable to any setting where SQL can be executed against the target database, which covers all standard Text to SQL benchmarks and the majority of deployed analytical workloads.

\bibliography{decosearch_v2/references}

@article{kahn1962topological,
    title={Topological sorting of large networks},
    author={Kahn, Arthur B.},
    journal={Communications of the ACM},
    volume={5},
    number={11},
    pages={558--562},
    year={1962},
    publisher={ACM}
}

@article{katsogiannis2023survey,
  title={A survey on deep learning approaches for text-to-{SQL}},
  author={Katsogiannis-Meimarakis, George and Koutrika, Georgia},
  journal={The VLDB Journal},
  volume={32},
  pages={905--936},
  year={2023},
  publisher={Springer}
}

@inproceedings{din,
    title={{DIN-SQL: Decomposed In-Context Learning of Text-to-SQL with Self-Correction}},
    author={Pourreza, Mohammadreza and Rafiei, Davood},
    booktitle={Advances in Neural Information Processing Systems (NeurIPS)},
    year={2023}
}

@inproceedings{chase_sql,
    title={{CHASE-SQL: Multi-Path Reasoning and Preference Optimized Candidate Selection in Text-to-SQL}},
    author={Pourreza, Mohammadreza and Chang, Hailong and Amin, Nima and others},
    booktitle={arXiv preprint arXiv:2410.01943},
    year={2024}
}

@inproceedings{chess_sql,
    title={{CHESS: Contextual Harnessing for Efficient SQL Synthesis}},
    author={Talaei, Shayan and Pourreza, Mohammadreza and Chang, Yu-Chen and others},
    booktitle={arXiv preprint arXiv:2405.16755},
    year={2024}
}

@inproceedings{bird_dataset,
    title={Can LLM Already Serve as A Database Interface? A BIg Bench for Large-Scale Database Grounded Text-to-SQLs},
    author={Li, Jinyang and Hui, Binyuan and Qu, Ge and others},
    booktitle={Advances in Neural Information Processing Systems},
    year={2023},
    volume={36}
}

@inproceedings{spider_dataset,
    title={Spider: A Large-Scale Human-Labeled Dataset for Complex and Cross-Domain Semantic Parsing and Text-to-SQL Task},
    author={Yu, Tao and Zhang, Rui and Yang, Kai and others},
    booktitle={Proceedings of the 2018 Conference on Empirical Methods in Natural Language Processing},
    year={2018}
}

@inproceedings{dail_sql,
    title={Text-to-SQL Empowered by Large Language Models: A Benchmark Evaluation},
    author={Gao, Dawei and Wang, Haibin and Li, Yaliang and others},
    booktitle={Proceedings of the VLDB Endowment},
    year={2024},
    volume={17},
    number={5}
}

@article{genasql,
    title={{Cheaper, Better, Faster, Stronger: Robust Text-to-SQL without Chain-of-Thought or Fine-Tuning}},
    author={D{\"o}nder, Yusuf Denizay and Hommel, Derek and Wen-Yi, Andrea W and Mimno, David and Jo, Eun Seo},
    journal={arXiv preprint arXiv:2505.14174},
    year={2025}
}

@article{omnisql,
    title={{OmniSQL}: Synthesizing High-quality Text-to-{SQL} Data at Scale},
    author={Li, Haoyang and Wu, Shang and Zhang, Xiaokang and Huang, Xinmei and Zhang, Jing and Jiang, Fuxin and Wang, Shuai and Zhang, Tieying and Chen, Jianjun and Shi, Rui and Chen, Hong and Li, Cuiping},
    journal={Proceedings of the VLDB Endowment},
    year={2025},
    note={arXiv:2503.02240}
    }

@inproceedings{deepeye_sql,
    title={{DeepEye-SQL: A Software-Engineering-Inspired Text-to-SQL Framework}},
    author={Boyan Li and Chong Chen and Zhujun Xue and Yinan Mei and Yuyu Luo},
    booktitle={ACM Special Interest Group on Management of Data (SIGMOD)},
    year={2026}
}

@article{agentar_scale_sql,
    title={{Agentar-Scale-SQL: Advancing Text-to-SQL through Orchestrated Test-Time Scaling}},
    author={Pengfei Wang and Baolin Sun and Xuemei Dong and Yaxun Dai and Hongwei Yuan and Mengdie Chu and Yingqi Gao and Xiang Qi and Peng Zhang and Ying Yan},
    journal={arXiv preprint arXiv:2509.24403},
    year={2025}
}

@article{xiyan_sql,
    title={{XiYan-SQL: A Novel Multi-Generator Framework For Text-to-SQL}},
    author={Yifu Liu and Yin Zhu and Yingqi Gao and Zhiling Luo and Xiaoxia Li and Xiaorong Shi and Yuntao Hong and Jinyang Gao and Yu Li and Bolin Ding and Jingren Zhou},
    journal={arXiv preprint arXiv:2507.04701},
    year={2026}
}

@inproceedings{sc,
    title={Self-Consistency Improves Chain of Thought Reasoning in Language Models},
    author={Wang, Xuezhi and Wei, Jason and Schuurmans, Dale and others},
    booktitle={International Conference on Learning Representations},
    year={2023}
}

@inproceedings{l2m,
    title={{Least-to-Most Prompting Enables Complex Reasoning in Large Language Models}},
    author={Zhou, Denny and Sch{\"a}rli, Nathanael and Hou, Le and others},
    booktitle={International Conference on Learning Representations (ICLR)},
    year={2023}
}

@inproceedings{selfrefine,
    title={Self-Refine: Iterative Refinement with Self-Feedback},
    author={Madaan, Aman and Tandon, Niket and Gupta, Pranjal and others},
    booktitle={Advances in Neural Information Processing Systems},
    year={2023},
    volume={36}
}

@inproceedings{critic,
    title={CRITIC: Large Language Models Can Self-Correct with Tool-Interactive Critiquing},
    author={Gou, Zhibin and Shao, Zhihong and Gong, Yeyun and others},
    booktitle={International Conference on Learning Representations},
    year={2024}
}

@inproceedings{igsql,
    title={{IGSQL: Database Schema Interaction Graph Based Neural Model for Context-Dependent Text-to-SQL Generation}},
    author={Cai, Yitao and Wan, Xiaojun},
    booktitle={2020 Conference on Empirical Methods in Natural Language Processing (EMNLP)},
    year={2020}
}

@inproceedings{ratsql,
    title={{RAT-SQL: Relation-Aware Schema Encoding and Linking for Text-to-SQL Parsers}},
    author={Wang, Bailin and Shin, Richard and Liu, Xiaodong and others},
    booktitle={Annual Meeting of the Association for Computational Linguistics (ACL)},
    year={2020}
}

@inproceedings{reimers-2019-sentence-bert,
    title={{Sentence-BERT: Sentence Embeddings using Siamese BERT-Networks}},
    author={Reimers, Nils and Gurevych, Iryna},
    booktitle={2019 Conference on Empirical Methods in Natural Language Processing and the 9th International Joint Conference on Natural Language Processing (EMNLP-IJCNLP)},
    year={2019}
}

@article{luo2025nl2sql,
  title={{Natural Language to SQL: State of the Art and Open Problems}},
  author={Luo, Yuyu and Li, Guoliang and Fan, Ju and Chai, Chengliang and Tang, Nan},
  journal={Proceedings of the VLDB Endowment},
  volume={18},
  number={12},
  pages={5466--33547131},
  year={2025}
}

@article{rajkumar2022evaluating,
  title={{Evaluating the Text-to-SQL Capabilities of Large Language Models}},
  author={Rajkumar, Nitarshan and Li, Raymond and Bahdanau, Dzmitry},
  journal={arXiv preprint arXiv:2204.00498},
  year={2022}
}

@article{li2024dawn,
  title={The Dawn of Natural Language to {SQL}: Are We Fully Ready?},
  author={Li, Boyan and Luo, Yuyu and Chai, Chengliang and Li, Guoliang and Tang, Nan},
  journal={Proceedings of the VLDB Endowment},
  volume={17},
  number={11},
  pages={3318--3331},
  year={2024}
}

@inproceedings{scholak2021picard,
  title={{PICARD: Parsing Incrementally for Constrained Auto-Regressive Decoding from Language Models}},
  author={Scholak, Torsten and Schucher, Nathan and Bahdanau, Dzmitry},
  booktitle={2021 Conference on Empirical Methods in Natural Language Processing (EMNLP)},
  year={2021}
}

@inproceedings{eyal2023qpl,
  title={{Semantic Decomposition of Question and SQL for Text-to-SQL Parsing}},
  author={Eyal, Ben and Mahabi, Moran and Haroche, Ophir and Bachar, Amir and Elhadad, Michael},
  booktitle={Findings of the Association for Computational Linguistics: EMNLP 2023},
  year={2023}
}

@inproceedings{wang2025macsql,
  title={{MAC-SQL: A Multi-Agent Collaborative Framework for Text-to-SQL}},
  author={Wang, Bing and Ren, Changyu and Yang, Jian and Liang, Xinnian and Bai, Jiaqi and Zhang, Qian-Wen and Yan, Zhao and Li, Zhoujun},
  booktitle={International Conference on Computational Linguistics (COLING 2025)},
  year={2025}
}

@inproceedings{chen2024selfdebug,
  title={{Teaching Large Language Models to Self-Debug}},
  author={Chen, Xinyun and Lin, Maxwell and Sch\"{a}rli, Nathanael and Zhou, Denny},
  booktitle={International Conference on Learning Representations (ICLR)},
  year={2024}
}

@inproceedings{askari2025magic,
  title={{MAGIC: Generating Self-Correction Guideline for In-Context Text-to-SQL}},
  author={Askari, Arian and Poelitz, Christian and Tang, Xinye},
  booktitle={AAAI Conference on Artificial Intelligence (AAAI)},
  year={2025}
}

@article{sun2023sqlpalm,
  title={{SQL-PaLM: Improved Large Language Model Adaptation for Text-to-SQL}},
  author={Sun, Ruoxi and Arik, Sercan and Muzio, Alex and Miculivicius, Lesly and Gundabathula, Satya and Yin, Pengcheng and Dai, Hanjun and Nakhost, Hootan and Sinha, Rajarishi and Wang, Zifeng and Pfister, Tomas},
  journal={arXiv preprint arXiv:2306.00739},
  year={2023}
}

@article{lee2024mcssql,
  title={{MCS-SQL: Leveraging Multiple Prompts and Multiple-Choice Selection for Text-to-SQL Generation}},
  author={Lee, Dongjun and Park, Choongwon and Kim, Jaehyuk and Park, Heesoo},
  journal={International Conference on Computational Linguistics (COLING)},
  year={2025}
}

@article{elliesql,
  title={{EllieSQL: Cost-Efficient Text-to-SQL with Complexity-Aware Routing}},
  author={Zhu, Yizhang and Jiang, Runzhi and Li, Boyan and Tang, Nan and Luo, Yuyu},
  journal={arXiv preprint arXiv:2503.22402},
  year={2025}
}

@article{query_and_conquer,
  title={{Query and Conquer: Execution-Guided SQL Generation}},
  author={Borchmann, {\L}ukasz and Wydmuch, Marek},
  journal={arXiv preprint arXiv:2503.24364},
  year={2025}
}

@inproceedings{spider2,
  title={{Spider 2.0: Evaluating Language Models on Real-World Enterprise Text-to-SQL Workflows}},
  author={Lei, Fangyu and Chen, Jixuan and Ye, Yuxiao and Cao, Ruisheng and Shin, Dongchan and Su, Hongjin and others},
  booktitle={International Conference on Learning Representations (ICLR)},
  year={2025}
}

@article{dea_sql,
  title={{Decomposition for Enhancing Attention: Improving {LLM}-based Text-to-{SQL} through Workflow Paradigm}},
  author={Xie, Yuanzhen and Jin, Xinzhou and Xie, Tao and Lin, MingXiong and Chen, Liang and others},
  journal={arXiv preprint arXiv:2402.10671},
  year={2024}
}

@article{rsl_sql,
  title={{RSL-SQL: Robust Schema Linking in Text-to-SQL Generation}},
  author={Cao, Zhenbiao and Zheng, Yuanlei and Fan, Zhihao and Zhang, Xiaojin and Chen, Wei and Bai, Xiang},
  journal={arXiv preprint arXiv:2411.00073},
  year={2024}
}

@article{excot,
  title={{ExCoT: Optimizing Reasoning for Text-to-SQL with Execution Feedback}},
  author={Zhai, Bohan and Xu, Canwen and He, Yuxiong and Yao, Zhewei},
  journal={arXiv preprint arXiv:2503.19988},
  year={2025}
}

@article{rethinking_agentic,
  title={{Rethinking Agentic Workflows: Evaluating Inference-Based Test-Time Scaling Strategies in Text-to-SQL Tasks}},
  author={Guo, Jiajing and Patel, Kenil and Ono, Jorge Piazentin and He, Wenbin and Ren, Liu},
  journal={arXiv preprint arXiv:2510.10885},
  year={2025}
}

\appendix

\clearpage
\newpage

\section{Detailed Algorithms}

\subsection{Schema Selector}
\label{app:schema_selector}

Algorithm~\ref{alg:schema_selector} details the \textsc{SchemaSelector} component that prunes the full database schema before any downstream LLM call.
The selector makes three sequential LLM calls: first to extract task-relevant keywords from the question and evidence, then to identify the necessary tables, and finally to pinpoint the required columns within those tables.
All LLM-returned names are remapped to their true schema counterparts via exact match followed by fuzzy match (FM), guarding against surface-level naming discrepancies in the LLM output.
\textsc{FuzzyMatch} finds the closest LLM-returned name to each true schema name by string similarity, returning $\varnothing$ if no sufficiently close match exists.
If the pruning procedure would produce an empty schema, the original schema is returned instead.

\begin{algorithm}[h!]
    \small
    \caption{\textsc{SchemaSelector}: LLM-Driven Schema Pruning}
    \label{alg:schema_selector}
    \begin{algorithmic}[1]
    \REQUIRE Question $Q$, full schema $\mathcal{S}$, evidence $E$
    \ENSURE Pruned schema $\mathcal{S}_{pruned}$, keyword list $K$

    \STATE \COMMENT{\textbf{Step 1: Keyword Extraction}}
    \STATE $K \leftarrow \text{LLM}_{\text{IR}}(Q, E)$
    \COMMENT{Extract keywords, entities, value mappings}

    \STATE \COMMENT{\textbf{Step 2: Table Selection}}
    \STATE $\mathcal{T}_{sel} \leftarrow \text{LLM}_{\text{TS}}(Q, E, K, \mathcal{S})$
    \COMMENT{Select relevant table names}

    \STATE \COMMENT{\textbf{Step 3: Column Selection}}
    \STATE $\mathcal{C}_{sel} \leftarrow \text{LLM}_{\text{CS}}(Q, E, K, \mathcal{T}_{sel})$
    \COMMENT{Select relevant columns per table}

    \STATE \COMMENT{\textbf{Step 4: Name Remapping \& Schema Assembly}}
    \STATE $\mathcal{S}_{pruned} \leftarrow \{\}$
    \FORALL{table $t \in \mathcal{S}$}
        \STATE $\hat{t} \leftarrow \text{FM}(t.\mathit{name},\; \mathcal{T}_{sel})$
        \IF{$\hat{t}$ found}
            \STATE $C_{t} \leftarrow \{c \in t.\mathit{cols} : \text{FM}(c.\mathit{name},\; \mathcal{C}_{sel}[\hat{t}])\}$
            \IF{$C_{t} \ne \emptyset$}
                \STATE $\mathcal{S}_{pruned}.\text{add}(t.\mathit{name},\; C_{t})$
            \ENDIF
        \ENDIF
    \ENDFOR

    \IF{$\mathcal{S}_{pruned} = \emptyset$}
        \RETURN $\mathcal{S},\; K$ \COMMENT{Fallback: return original schema}
    \ENDIF
    \RETURN $\mathcal{S}_{pruned},\; K$
    \end{algorithmic}
\end{algorithm}

\subsection{Topological Execution Order}
\label{app:topo_sort}

Algorithm~\ref{alg:topo_sort} computes the node execution order from the decomposition DAG using Kahn's algorithm~\cite{kahn1962topological}.
In the DAG $G=(V,E)$, a directed edge $(u,v)$ indicates that node $v$ consumes the result of node $u$, so $u$ must execute before $v$.
Leaf nodes (in-degree zero) have no upstream dependencies and are scheduled first; a cycle detection check guards against malformed LLM outputs.

\begin{algorithm}[h!]
    \small
    \caption{\textsc{TopologicalSort}: Dependency-Aware Execution Ordering}
    \label{alg:topo_sort}
    \begin{algorithmic}[1]
    \REQUIRE DAG $G = (V, E)$; edge $(u,v)$ means $v$ depends on $u$
    \ENSURE Execution queue $\mathit{Queue}$ with all dependencies before dependents
    \STATE $\mathit{depCount}[v] \leftarrow |\{u : (u,v) \in E\}|$ \quad for all $v \in V$
    \STATE $\mathit{Ready} \leftarrow \{v \in V : \mathit{depCount}[v] = 0\}$ \COMMENT{Leaf nodes}
    \STATE $\mathit{Queue} \leftarrow [\;]$
    \WHILE{$\mathit{Ready}$ is not empty}
        \STATE $n \leftarrow \mathit{Ready}.\text{pop}()$
        \STATE $\mathit{Queue}.\text{append}(n)$
        \FORALL{dependent $d$ such that $(n, d) \in E$}
            \STATE $\mathit{depCount}[d] \leftarrow \mathit{depCount}[d] - 1$
            \IF{$\mathit{depCount}[d] = 0$}
                \STATE $\mathit{Ready}.\text{add}(d)$ \COMMENT{All of $d$'s dependencies resolved}
            \ENDIF
        \ENDFOR
    \ENDWHILE
    \IF{$|\mathit{Queue}| < |V|$}
        \STATE \textbf{raise} \textsc{CycleDetectedError}
    \ENDIF
    \RETURN $\mathit{Queue}$
    \end{algorithmic}
\end{algorithm}

\subsection{Dependency Placeholder Resolution}
\label{app:placeholder_resolution}

When a downstream node depends on the result of an upstream node $u$ via the \texttt{@[$u$]} placeholder, DecoSearch must inject that result into the downstream node's sub-question and SQL prompt before generation begins.
The injection strategy is chosen based on the cardinality of node $u$'s result set.

\paragraph{Small result sets ($|\mathcal{R}_u| \leq \theta$).}
If the result contains at most $\theta = 100$ rows, the values are serialized directly as a SQL value list.
For example, \texttt{@[1]} in the sub-question text is replaced with the literal tuple \texttt{('v1', 'v2', \ldots)}, which can appear directly inside an \texttt{IN (\ldots)} clause in the generated SQL.

\paragraph{Large result sets ($|\mathcal{R}_u| > \theta$).}
When the result exceeds $\theta$ rows, inlining the values would produce an unwieldy prompt and an oversized SQL literal.
Instead, DecoSearch materializes the result into a \emph{SQLite temporary table} using:
\begin{center}
\texttt{CREATE TEMPORARY TABLE temp\_ds\_node\_$u$ AS $\langle$SQL$_u\rangle$}
\end{center}
The temporary table \texttt{temp\_ds\_node\_$u$} is then referenced by name wherever \texttt{@[$u$]} appears in downstream SQL, and a schema description comment is prepended to the downstream node's prompt so the LLM knows the table exists and can query it like any ordinary table.
The temporary table persists for the lifetime of the database connection and is therefore available to all subsequent nodes in the same execution run.

\noindent \paragraph{Fallback.}
If temporary-table creation is disabled or the result is empty, values are truncated to a adjustable limit ($200$ rows by default) with a comment appended to the SQL noting the truncation.

\subsection{Node-Level Solve}
\label{app:solve_node}

Algorithm~\ref{alg:solve_node} details \textsc{SolveNode}, invoked once per popped queue node from the main pipeline (Algorithm~\ref{alg:sc_algorithm}). It implements query-level generation (\S\ref{sec:generation}), a single query-level repair attempt via the SQL Refiner (\S\ref{sec:sql_refiner}), and, on persistent failure, escalation to the Topology Refiner (Algorithm~\ref{alg:topology_refiner}) subject to the shared per-query budget $B$. If the budget is exhausted or the revision is rejected, it accepts the best available result for $n$ and returns it unchanged, matching the graceful-degradation behavior described in \S\ref{sec:topology_refiner}.

\begin{algorithm}[h!]
    \small
    \caption{\textsc{SolveNode}: Generate, Repair, and Materialize a Sub-Question}
    \label{alg:solve_node}
    \begin{algorithmic}[1]
    \REQUIRE Node $n$, DAG $G$, partial results $Results$, pruned schema $\mathcal{S}_{pruned}$, database $\mathcal{D}$, remaining topology-refinement budget $B$
    \ENSURE Result for $n$ (or \text{NONE} if $G$ is revised and $n$ must be re-solved under the new graph), and $G$ (unchanged, or revised)
    \STATE $Q_{sub} \leftarrow \text{Substitute}(n.question, Results)$
    \STATE $\hat{Q}_{sub} \leftarrow \text{SQLGenerator}(Q_{sub}, \mathcal{S}_{pruned})$
    \STATE $R_{sub}, \text{Error} \leftarrow \text{Execute}(\mathcal{D}, \hat{Q}_{sub})$
    \IF{$\text{Error} \ne \text{NONE}$ \OR $R_{sub} = \emptyset$}
        \STATE \COMMENT{Query-level repair}
        \STATE $\hat{Q}_{sub} \leftarrow \text{SQLRefiner}(Q_{sub}, \hat{Q}_{sub}, \text{Error})$
        \STATE $R_{sub}, \text{Error} \leftarrow \text{Execute}(\mathcal{D}, \hat{Q}_{sub})$
    \ENDIF
    \IF{$\text{Error} \ne \text{NONE}$ \OR $R_{sub} = \emptyset$}
        \IF{$B > 0$}
            \STATE \COMMENT{Plan-level repair}
            \STATE $G' \leftarrow \text{TopologyRefiner}(G, n, \text{Error})$
            \IF{$G'$ is revised}
                \RETURN $\text{NONE}, G'$ \COMMENT{$B \leftarrow B-1$; caller resumes from $G'$}
            \ENDIF
        \ENDIF
        \STATE \COMMENT{Budget exhausted or no viable revision: accept best available}
    \ENDIF
    \RETURN $\text{Materialize}(\mathcal{D}, \hat{Q}_{sub}, R_{sub}), G$
    \end{algorithmic}
\end{algorithm}

\subsection{Topology Refiner}
\label{app:topology_refiner}

Algorithm~\ref{alg:topology_refiner} details \textsc{TopologyRefiner}, invoked from \textsc{SolveNode} (Algorithm~\ref{alg:solve_node}) when a node still fails after query-level repair by the SQL Refiner and the refinement budget $B$ has not been exhausted.
The refiner serializes the current graph to JSON, constructs a prompt that includes the failing node's sub-question and the execution error, and asks the LLM to produce a structurally revised DAG.
The caller rejects the revision if it is isomorphic to the original (no structural change) or contains a cycle, and falls back to the original graph in either case.

\begin{algorithm}[h!]
    \small
    \caption{\textsc{TopologyRefiner}: LLM-Driven DAG Revision}
    \label{alg:topology_refiner}
    \begin{algorithmic}[1]
    \REQUIRE DAG $G$, failing node $n^*$, failure reason $\rho$, question $Q$, evidence $E$
    \ENSURE Revised DAG $G'$ (or $G$ if revision is invalid)
    \STATE $J \leftarrow \text{Serialize}(G)$ \COMMENT{Graph $\to$ JSON}
    \STATE $\mathit{prompt} \leftarrow \text{BuildPrompt}(Q,\, E,\, J,\, n^*,\, \rho)$
    \STATE $\mathit{response} \leftarrow \text{LLM}(\mathit{prompt})$
    \STATE $J' \leftarrow \text{ExtractJSON}(\mathit{response})$
    \IF{$J'$ is valid JSON}
        \STATE $G' \leftarrow \text{Deserialize}(J')$
        \IF{$G'$ is isomorphic to $G$}
            \RETURN $G$ \COMMENT{No structural change; skip}
        \ENDIF
        \IF{$G'$ contains a cycle}
            \RETURN $G$ \COMMENT{Reject invalid DAG}
        \ENDIF
        \RETURN $G'$
    \ELSE
        \RETURN $G$ \COMMENT{Parse failure; retain original}
    \ENDIF
    \end{algorithmic}
\end{algorithm}

\section{Detailed Experiment Setup}
\label{app:hyperparams}
All experiments on both BIRD and Spider use the same configuration unless otherwise stated.
DeepSeek-V3 (\texttt{deepseek-chat}) serves as the backbone for every LLM role in the pipeline, and a global random seed of $42$ is fixed across all API calls to ensure reproducibility.
The judger and the direct SQL generator both run at temperature $T = 0$ to produce deterministic outputs, while the decomposer runs at $T = 0.1$ to allow for slightly varied but still constrained decomposition strategies.
The RAG-based decomposer retrieves $k = 3$ similar examples from the training knowledge base using \texttt{all-MiniLM-L6-v2} embeddings to construct few-shot decomposition prompts.
The SQL Refiner is invoked at most once per DAG node, on execution errors or empty results, before the topology refiner is considered.
If execution still fails at any DAG node, the topology refiner is invoked for up to $B = 3$ revision attempts before falling back to the best available result.
The schema selector runs a three-stage LLM pipeline (keyword extraction, table selection, column selection) to prune the full database schema before any downstream call; pruning is adaptive with no hard cap on the number of retained tables, and fuzzy matching is applied to guard against naming discrepancies in the LLM output.

\section{Prompts}
\label{app:prompts}

We report the core prompt templates used by each pipeline component.
Slots filled dynamically at runtime are shown in \textcolor{blue!60!black}{\texttt{[SLOT]}}.

\subsection{Judger Prompt}
\label{app:prompt_judger}





\begin{promptbox}{Judger Prompt}
\small\ttfamily
\textbf{System:} You are an expert SQL architect and query planner. Your job is to analyze a natural language question and a database schema to decide if the question should be decomposed into multiple sub-questions or if it can be solved reliably with a single direct SQL query.

\medskip
\textbf{STRATEGY: PREFER DIRECT SQL} \\
Empirical analysis shows that our Direct SQL path is highly accurate. Decomposition should be used ONLY as a "Heavy Lifting" tool for logic that cannot be reliably expressed in a single pass.

\medskip
\textbf{1. When to choose Direct SQL (needs\_decomposition: false):}
\begin{itemize}[nosep,leftmargin=1.2em]
  \item Standard Relational Lookups: Even with 2--4 joins, a single query is usually best.
  \item Simple Filters \& Mappings: Using values from the Hint to filter columns.
  \item Standard Aggregations: SUM, AVG, COUNT, MAX/MIN per group.
  \item Basic Subqueries: \texttt{WHERE id IN (SELECT ...)} or simple CTEs.
\end{itemize}

\medskip
\textbf{2. When to choose Decomposition (needs\_decomposition: true):}
\begin{itemize}[nosep,leftmargin=1.2em]
  \item \textbf{Conflicting Aggregation Grains:} Asking for a detail and a scalar at once (e.g., ``List all students AND the average score'').
  \item \textbf{Extreme Join Complexity:} Joining $\geq$5 tables where the path is highly ambiguous.
  \item \textbf{Complex Multi-Phase Reasoning:} When Step~2 depends on a mathematical result from Step~1 that must be calculated first (e.g., ``Find the top earner, then find all their transactions'').
  \item \textbf{Evidence-Driven Fork:} If the Hint specifies a formula or complex mapping that makes a single query extremely nested and error-prone.
\end{itemize}

\medskip
\textbf{Input:}\\
Database Schema: \textcolor{blue!60!black}{\texttt{\{schema\}}}\\
Natural Language Question: \textcolor{blue!60!black}{\texttt{\{question\}}}\\
Task: Your task is to analyze the question and schema. Decide if the question NEEDS to be decomposed.

\medskip
\textbf{Output:} Respond with a JSON object. Only output the JSON object. Do not include any other text or markdown formatting.\\
\texttt{\{"needs\_decomposition": true/false, "reasoning": "A brief explanation of your decision based on the criteria (simplicity, JOINs, logical steps)."\} }

\medskip
Now, take a deep breath, and think step by step to make your decision.
\end{promptbox}

\subsection{Decomposer Prompt}
\label{app:prompt_decomposer}

\begin{promptbox}{Decomposition Planner Prompt}
\small\ttfamily
\textbf{System:} You are a specialized SQL Architect and Query Planner. Your task is to decompose a complex natural language question into a logically sound dependency graph (DAG) of simpler sub-questions.

\medskip
\textbf{MISSION:} Break the question into atomic, executable steps where each step produces a result required by the next.

\medskip
\textbf{STRICT DECOMPOSITION RULES:}
\begin{itemize}[nosep,leftmargin=1.2em]
  \item \textbf{1. MANDATORY IDENTIFIER PASSING:} If Step~A depends on Step~B, Step~B MUST return a specific identifier (like an ID, Code, or Name) that Step~A can use for filtering. Example: ``Find the IDs of...'' followed by ``Find the details for IDs in @[ID]''.
  \item \textbf{2. PLACEHOLDER SYNTAX:} Use \texttt{@[node\_id]} to reference the result of a previous step. 
  \item \textbf{3. EVIDENCE IS THE MAP:} BIRD Evidence often contains formulas or value mappings. If the evidence says ```Active' means status=1'', create a sub-question that handles that specific filter or calculation.
  \item \textbf{4. NO FRAGMENTATION:} Do not create a new node for every single clause. Only decompose if the result of one query is logically required to build the next (e.g., Finding a Max/Min, or finding a set of IDs to filter a different table).
  \item \textbf{5. ROOT NODE PRECISION:} The root node (ID 0) must be the final transformation. It should return the exact column(s) requested in the original question using the results from its children.
\end{itemize}

\medskip
Use the provided successful examples below to understand the correct decomposition format and style.

\medskip
\textcolor{blue!60!black}{\texttt{\{examples\_prompt\_str\}}}

\medskip
\textbf{Input:}\\
Database Schema: \textcolor{blue!60!black}{\texttt{\{schema\}}}\\
Evidence (Hint): \textcolor{blue!60!black}{\texttt{\{evidence\}}} \textit{(Optional)} \\
Question: \textcolor{blue!60!black}{\texttt{\{question\}}}

\medskip
\textbf{Output Format (JSON Only):}
\begin{itemize}[nosep,leftmargin=1.2em]
  \item Output must be a single JSON object with \texttt{nodes} and \texttt{edges}.
  \item Nodes: \texttt{id} (integer) and \texttt{question} (string).
  \item Edges: \texttt{[parent\_id, child\_id]}, where parent depends on child.
  \item The graph must be a DAG, flowing from leaves to root.
\end{itemize}

\medskip
Output only the valid JSON representation of the graph.
\end{promptbox}

\subsection{SQL Generator Prompt}
\label{app:prompt_generator}

\begin{promptbox}{SQL Generator Prompt - BIRD}
\small\ttfamily
\textbf{System:} You are an expert SQL generator. Below, you are presented with a database schema and a question. Your task is to read the schema, understand the question, and generate a valid \textcolor{blue!60!black}{\texttt{\{self.db\_type\}}} query to answer the question.

\medskip
\textbf{Input:}\\
Database Schema: \textcolor{blue!60!black}{\texttt{\{schema\_str\}}}\\
Question: \textcolor{blue!60!black}{\texttt{\{sub\_question\}}}\\
Hint (Evidence): \textcolor{blue!60!black}{\texttt{\{evidence\}}}

\medskip
\textbf{MANDATORY BIRD EXECUTION RULES:}
\begin{itemize}[nosep,leftmargin=1.2em]
  \item \textbf{1. The Hint is GROUND TRUTH:} The Hint (Evidence) is a strict requirement. If the hint provides a mapping (e.g., ``'A''' for ``'Active''') or a specific filter value, you MUST use it exactly as specified. Do not deviate from the hint.
  \item \textbf{2. Case-Insensitivity:} SQLite is case-sensitive for string comparisons. Use \texttt{LOWER()} on columns and filter values (e.g., \texttt{LOWER(city) = LOWER('London')}) or \texttt{LIKE} for all string filters unless the hint specifies otherwise.
  \item \textbf{3. Joins:} Use explicit \texttt{JOIN ... ON} syntax.
  \item \textbf{4. Name vs ID:} Always prefer using the name column over the id column unless the hint specifies otherwise.
  \item \textbf{5. Null Handling:} Be mindful of NULL values in your logic, especially in aggregations and comparisons.
  \item \textbf{6. Final Check:} Ensure your query returns all requested information and nothing extra.
\end{itemize}

\medskip
\textbf{Output:} Please respond with a JSON object structured as follows:\\
\texttt{\{"SQL": "Your SQL query in a single string."\}}

\medskip
Take a deep breath and think step by step to find the correct SQL query.
\end{promptbox}

\begin{promptbox}{SQL Generator Prompt - Spider}
\small\ttfamily
\textbf{System:} You are a data science expert. Below, you are presented with a database schema and a question. Your task is to read the schema, understand the question, and generate a valid \textcolor{blue!60!black}{\texttt{\{self.db\_type\}}} query to answer the question. Before generating the final SQL query think step by step on how to write the query.

\medskip
\textbf{Input:}\\
Database Schema: \textcolor{blue!60!black}{\texttt{\{schema\_str\}}}\\
Question: \textcolor{blue!60!black}{\texttt{\{sub\_question\}}}\\
Hint: \textcolor{blue!60!black}{\texttt{\{evidence\}}}

\medskip
This schema offers an in-depth description of the database's architecture, detailing tables, columns, primary keys, foreign keys, and any pertinent information regarding relationships or constraints. Special attention should be given to the examples listed beside each column, as they directly hint at which columns are relevant to our query.

\medskip
\textbf{Database instructions:}
\begin{itemize}[nosep,leftmargin=1.2em]
  \item Make sure you only output the information that is asked in the question. If the question asks for a specific column, make sure to only include that column in the SELECT clause, nothing more.
  \item Predicted query should return all of the information asked in the question without any missing or extra information.
\end{itemize}

\medskip
Priority should be given to columns that have been explicitly matched with examples relevant to the question's context.

\medskip
\textbf{Output:} Please respond with a JSON object structured as follows:\\
\texttt{\{} \\
\texttt{~~"chain\_of\_thought\_reasoning": "Your thought process...",} \\
\texttt{~~"SQL": "Your SQL query in a single string."} \\
\texttt{\}}

\medskip
Take a deep breath and think step by step to find the correct \textcolor{blue!60!black}{\texttt{\{self.db\_type\}}} SQL query.
\end{promptbox}

\subsection{SQL Refiner Prompt}
\label{app:prompt_refiner}

\begin{promptbox}{SQL Refiner Prompt}
\small\ttfamily
\textbf{System:} You are an expert SQL debugger. Your task is to fix a SQL query based on a history of failed attempts and their execution results. Your objective is to make sure the query follows the database admin instructions and uses the correct conditions.

\medskip
\textbf{Input:}\\
Original Question: \textcolor{blue!60!black}{\texttt{\{sub\_question\}}}\\
Database Schema: \textcolor{blue!60!black}{\texttt{\{schema\_str\}}}\\
Evidence (Hint): \textcolor{blue!60!black}{\texttt{\{evidence\_prompt\}}}

\medskip
\textbf{DATABASE ADMIN INSTRUCTIONS:}
\begin{itemize}[nosep,leftmargin=1.2em]
  \item \textbf{1. Output Scope:} Only output the information that is asked in the question. If the question asks for a specific column, include only that column in the \texttt{SELECT} clause, nothing more.
  \item \textbf{2. Completeness:} The predicted query should return all of the information asked in the question, without any missing or extra information.
  \item \textbf{3. Counting Grain:} Before writing \texttt{COUNT(...)}, identify which column is the actual entity being counted from the schema's primary/foreign keys. If a join can produce multiple rows per entity, use \texttt{COUNT(DISTINCT <entity\_id>)} rather than a bare \texttt{COUNT(*)}/\texttt{COUNT(column)}, which would overcount.
  \item \textbf{4. Ties on Extremes:} When the question asks for ``the X with the most/least/highest/lowest Y'' and does not explicitly ask for a single top-1 result, prefer \texttt{WHERE Y = (SELECT MAX(Y) ...)} or a \texttt{HAVING} clause over \texttt{ORDER BY Y DESC LIMIT 1}, which silently drops legitimate ties. Use \texttt{LIMIT 1} only when the question explicitly asks for exactly one row.
  \item \textbf{5. Mixed Aggregation Grains:} If the question asks for a per-group value AND a separate overall value in the same answer, compute each aggregate in its own subquery/CTE at its own grain; do not produce both from a single \texttt{GROUP BY}.
  \item \textbf{6. Prior-Step Results:} The question may reference a temporary table \texttt{temp\_ds\_node\_<N>}, the already-computed result of an earlier step. Treat its contents as fixed and correct; \texttt{JOIN} or subquery against it directly rather than re-deriving, re-filtering, or hardcoding literal values that duplicate it.
\end{itemize}

\medskip
\textbf{Attempt History:} You have already made \textcolor{blue!60!black}{\texttt{\{n\_attempts\}}} attempt(s) that resulted in an error or an incorrect answer. Review the history of these attempts to understand what went wrong.\\
\textcolor{blue!60!black}{\texttt{\{history\_str\}}}

\medskip
\textbf{Task:} Based on the full history of failures, provide a new, corrected SQL query that addresses all the issues identified above. Do not include a trailing semicolon.

\medskip
\textbf{Output:} Corrected SQL query.
\end{promptbox}

\subsection{Topology Refiner Prompt}
\label{app:prompt_topology}

\begin{promptbox}{Topology Refiner Prompt}
\small\ttfamily
\textbf{System:} You are an expert workflow planner. Your task is to revise a task decomposition graph that has failed during execution.

\medskip
\textbf{Input:}\\
Original User Question: \textcolor{blue!60!black}{\texttt{\{question\}}}\\
Current Decomposition Graph (JSON): \textcolor{blue!60!black}{\texttt{\{graph\_json\_str\}}}\\
Execution Failure:
\begin{itemize}[nosep,leftmargin=1.2em]
  \item The plan failed at \textbf{Node \textcolor{blue!60!black}{\texttt{\{failed\_node\_id\}}}}.
  \item Sub-question for this node: \textcolor{blue!60!black}{\texttt{"\{sub-question\}"}}
  \item Reason for failure: \textcolor{blue!60!black}{\texttt{"\{failure\_reason\}"}}
\end{itemize}

\medskip
\textbf{Your Task:} Analyze the failure and the overall graph, then propose a corrected version of the graph in JSON format.

\medskip
\textbf{Revision Rules:}
\begin{itemize}[nosep,leftmargin=1.2em]
  \item \textbf{1. Localize Changes:} Make the minimum necessary changes to fix the issue. Focus on the area around the failed node.
  \item \textbf{2. Allowed Operations:}
  \begin{itemize}[nosep,leftmargin=1.2em]
    \item \textit{Rephrase a node:} Modify the 'question' of a node to be clearer or more specific.
    \item \textit{Decompose a node:} Replace a single failed node with two or more simpler nodes. Ensure you correctly wire the new dependencies.
    \item \textit{Add a node:} Add a new intermediate step if a prerequisite was missing.
    \item \textit{Rewire edges:} Change the dependencies (\texttt{edges}) if the data flow was incorrect.
  \end{itemize}
  \item \textbf{3. Preserve Completed Nodes:} Do NOT modify nodes that have already executed successfully (i.e., nodes that are dependencies of the failed node).
  \item \textbf{4. Maintain Graph Integrity:} The output must be a valid Directed Acyclic Graph (DAG). Do not introduce cycles.
  \item \textbf{5. Output Format:} Your output MUST be only the complete, revised graph in the same JSON format as the input. Do not include any other text, explanations, or markdown formatting.
\end{itemize}

\medskip
\textbf{Output:} Revised Graph (JSON only):\\
\texttt{\{} \\
\texttt{~~"nodes": [...],} \\
\texttt{~~"edges": [...]} \\
\texttt{\}}
\end{promptbox}

\section{Case Studies}
\label{app:case_studies}

We present two worked examples from the BIRD development set illustrating how DecoSearch decomposes complex questions into a dependency DAG and solves each sub-question in topological order.
Both examples were solved correctly by DecoSearch while generating a direct SQL attempt failed to produce the right answer.

\subsection{Example 1: Sequential Dependency Chain}
\label{app:case1}

\begin{promptbox}{Question \& Context --- \texttt{formula\_1} database}
\small\ttfamily
\textbf{NLQ:} How many accidents did the driver who had the highest number of accidents in the Canadian Grand Prix have?

\medskip
\textbf{Evidence:} \textit{accidents} $\equiv$ rows where \texttt{statusId = 3};\quad
\textit{Canadian Grand Prix} $\equiv$ \texttt{races.name = `Canadian Grand Prix'}

\medskip
\textbf{Judger decision:} \textbf{Decompose.}
The question requires a two-phase aggregation: first materialise the set of Canadian GP race IDs, then identify the most-accident driver in those races, and finally count their accidents. Expressing this as a single \texttt{MAX(COUNT(\ldots))} subquery is error-prone in one-shot generation.
\end{promptbox}

\noindent
\textbf{DAG} (3 nodes; execution order $1 \to 2 \to 0$):

\smallskip
\begin{center}
\begin{tikzpicture}[
  >=Stealth,
  leafnode/.style={circle, draw, rounded corners=3pt, fill=blue!10,
                   text width=1.9cm, minimum height=0.85cm, align=center, font=\small},
  midnode/.style= {circle, draw, rounded corners=3pt, fill=gray!8,
                   text width=1.9cm, minimum height=0.85cm, align=center, font=\small},
  rootnode/.style={circle, draw, rounded corners=3pt, fill=green!12,
                   text width=1.7cm, minimum height=0.85cm, align=center, font=\small},
  arr/.style={->, thick, gray!55!black}
]
  \node[leafnode] (n1) at ( 0.0,  1.0)
      {\textbf{Node 1}\\\footnotesize raceId lookup};
  \node[midnode]  (n2) at (-1.8, -2.0)
      {\textbf{Node 2}\\\footnotesize top driver\\\footnotesize\texttt{@[1]}};
  \node[rootnode] (n0) at ( 1.8, -2.0)
      {\textbf{Node 0}\\\footnotesize count accidents\\\footnotesize\texttt{@[1]}, \texttt{@[2]}};
  \draw[arr] (n1) -- (n2);
  \draw[arr] (n1) -- (n0);
  \draw[arr] (n2) -- (n0);
\end{tikzpicture}\\[4pt]
{\footnotesize\textit{\colorbox{blue!10}{leaf}\enspace
                      \colorbox{gray!8}{intermediate}\enspace
                      \colorbox{green!12}{root}}}
\end{center}

\smallskip
\begin{center}
\small
\begin{tabular}{@{}clp{0.55\linewidth}@{}}
\toprule
\textbf{Node} & \textbf{Deps.} & \textbf{Sub-question} \\
\midrule
1 & --- & Find the \texttt{raceId}(s) for the race named \emph{`Canadian Grand Prix'}. \\
2 & \texttt{@[1]} & Find the \texttt{driverId} with the most accidents (\texttt{statusId=3}) among races \texttt{@[1]}. \\
0 & \texttt{@[1]}, \texttt{@[2]} & Count accidents for driver \texttt{@[2]} across races \texttt{@[1]}. \\
\bottomrule
\end{tabular}
\end{center}

\smallskip

\begin{promptbox}{Node 1 (leaf) \textrm{---} no dependencies}
\small\ttfamily
SELECT raceId FROM races\\
WHERE name = 'Canadian Grand Prix'
\end{promptbox}

\begin{promptbox}{Node 2 \textrm{---} uses \texttt{@[1]}}
\small\ttfamily
SELECT driverId FROM results\\
WHERE statusId = 3 AND raceId IN @[1]\\
GROUP BY driverId ORDER BY COUNT(*) DESC LIMIT 1
\end{promptbox}

\begin{promptbox}{Node 0 (root) \textrm{---} uses \texttt{@[1]}, \texttt{@[2]}}
\small\ttfamily
SELECT COUNT(*) FROM results\\
WHERE driverId = @[2] AND statusId = 3 AND raceId IN @[1]
\end{promptbox}

\noindent\textbf{Answer:} \texttt{2} \quad \textcolor{green!60!black}{\checkmark}~Matches gold.

\medskip\noindent
\textit{Why decomposition helps:} 
Direct queries for this pattern demand a nested \texttt{MAX(COUNT(...))} structure, a complexity that often causes LLMs to hallucinate or mistake it for a flat aggregation. By materializing the driver ID as an intermediate result, the task is simplified into basic \texttt{GROUP BY} operations or filters, which the model processes reliably.
\subsection{Example 2: Parallel Branches with Aggregation}
\label{app:case2}

\begin{promptbox}{Question \& Context --- \texttt{thrombosis\_prediction} database}
\small\ttfamily
\textbf{NLQ:} What is the ratio of male to female patients among all those with abnormal uric acid counts?

\medskip
\textbf{Evidence:} \textit{male} $\equiv$ \texttt{SEX=`M'}; \textit{female} $\equiv$ \texttt{SEX=`F'};
\textit{abnormal UA} $\equiv$ \texttt{UA}$\leq$\texttt{8.0} for males, \texttt{UA}$\leq$\texttt{6.5} for females;
$\text{ratio} = \texttt{DIVIDE(count\_M, count\_F)}$

\medskip
\textbf{Judger decision:} \textbf{Decompose.}
The ratio depends on two separate conditional counts with sex-specific thresholds.
Decomposition computes each count independently, then combines them in the root node, avoiding the compound \texttt{CASE}-within-\texttt{CASE} logic that often produces off-by-one errors in direct generation.
\end{promptbox}

\noindent
\textbf{DAG} (4 nodes; execution order $3 \to 1, 2 \to 0$; diamond topology):

\smallskip
\begin{center}
\begin{tikzpicture}[
  >=Stealth,
  leafnode/.style={circle, draw, rounded corners=3pt, fill=blue!10,
                   text width=1.5cm, minimum height=0.85cm, align=center, font=\small},
  midnode/.style= {circle, draw, rounded corners=3pt, fill=gray!8,
                   text width=1.5cm, minimum height=0.85cm, align=center, font=\small},
  rootnode/.style={circle, draw, rounded corners=3pt, fill=green!12,
                   text width=1.8cm, minimum height=0.85cm, align=center, font=\small},
  arr/.style={->, thick, gray!55!black}
]
  \node[leafnode] (n3) at ( 0.0,  0.8)
      {\textbf{Node 3}\\\footnotesize abnormal UA\\\footnotesize patient IDs};
  \node[midnode]  (n1) at (-1.8, -2.0)
      {\textbf{Node 1}\\\footnotesize count males\\\footnotesize\texttt{@[3]}};
  \node[midnode]  (n2) at ( 1.8, -2.0)
      {\textbf{Node 2}\\\footnotesize count females\\\footnotesize\texttt{@[3]}};
  \node[rootnode] (n0) at ( 0.0, -5.0)
      {\textbf{Node 0}\\\footnotesize M:F ratio\\\footnotesize\texttt{@[1]}, \texttt{@[2]}};
  \draw[arr] (n3) -- (n1);
  \draw[arr] (n3) -- (n2);
  \draw[arr] (n1) -- (n0);
  \draw[arr] (n2) -- (n0);
\end{tikzpicture}\\[4pt]
{\footnotesize\textit{\colorbox{blue!10}{leaf}\enspace
                      \colorbox{gray!8}{intermediate}\enspace
                      \colorbox{green!12}{root}}}
\end{center}

\smallskip
\begin{center}
\small
\begin{tabular}{@{}clp{0.55\linewidth}@{}}
\toprule
\textbf{Node} & \textbf{Deps.} & \textbf{Sub-question} \\
\midrule
3 & --- & Find all patient IDs where UA is abnormal: (\texttt{SEX='M'} \& \texttt{UA}$\leq$\texttt{8.0}) \texttt{OR} (\texttt{SEX='F'} \& \texttt{UA}$\leq$\texttt{6.5}). \\
1 & \texttt{@[3]} & Count distinct male patients (\texttt{SEX='M'}) from \texttt{@[3]}. \\
2 & \texttt{@[3]} & Count distinct female patients (\texttt{SEX='F'}) from \texttt{@[3]}. \\
0 & \texttt{@[1]}, \texttt{@[2]} & Compute \texttt{CAST(@[1] AS REAL) / @[2]}. \\
\bottomrule
\end{tabular}
\end{center}

\smallskip
\begin{promptbox}{Node 3 (leaf) \textrm{---} no dependencies}
\small\ttfamily
SELECT DISTINCT l.ID FROM Laboratory l\\
JOIN Patient p ON l.ID = p.ID\\
WHERE (p.SEX = 'M' AND l.UA <= 8.0)\\
\phantom{WHERE }OR (p.SEX = 'F' AND l.UA <= 6.5)
\end{promptbox}

\begin{promptbox}{Node 1 \textrm{---} uses \texttt{@[3]}}
\small\ttfamily
SELECT COUNT(DISTINCT p.ID) FROM Patient p\\
WHERE p.ID IN @[3] AND p.SEX = 'M'
\end{promptbox}

\begin{promptbox}{Node 2 \textrm{---} uses \texttt{@[3]}}
\small\ttfamily
SELECT COUNT(DISTINCT p.ID) FROM Patient p\\
WHERE p.ID IN @[3] AND p.SEX = 'F'
\end{promptbox}

\begin{promptbox}{Node 0 (root) \textrm{---} uses \texttt{@[1]}, \texttt{@[2]}}
\small\ttfamily
SELECT CAST(@[1] AS REAL) / NULLIF(@[2], 0) AS ratio
\end{promptbox}

\noindent\textbf{Answer:} \texttt{0.2057} \quad \textcolor{green!60!black}{\checkmark}~Matches gold.

\medskip\noindent
\textit{Why decomposition helps:} The sex-specific abnormality thresholds make it natural to separate the population filtering (Node~3) from the sex-conditional counts (Nodes~1 and~2).
A direct query must interleave both concerns in a single \texttt{CASE} expression, a pattern where LLMs frequently apply the wrong threshold to the wrong gender.
The diamond topology here shows that DecoSearch is not limited to linear chains: two sibling nodes can share a common dependency and produce independent partial results that are combined at the root.

\section{LLM Usage}  
In this paper, we leverage LLMs, including ChatGPT and Gemini, to refine sentence-level writing.

\end{document}